\documentclass[letterpaper, 10 pt, conference]{ieeeconf}  

\IEEEoverridecommandlockouts                              

\usepackage{amsmath,bm}
\usepackage{amsfonts,amssymb}
\usepackage{booktabs}
\usepackage{graphicx} 
\usepackage{float}
\usepackage{subfigure}
\usepackage{bm}
\usepackage{color}
\usepackage{multirow}

\newcommand{\vts}{\mathrm{T}}
\newcommand{\bv}[1]{{\bm {#1}}}
\newcommand{\ba}[1]{{\bf {#1}}}
\newcommand{\beq}{\begin{equation}}
\newcommand{\eeq}{\end{equation}}
\newcommand{\bmat}{\begin{bmatrix}}
	\newcommand{\emat}{\end{bmatrix}}
\newcommand{\Eq}[1]{(\ref{#1})}
\newcommand{\Fig}[1]{Fig. \ref{#1}}
\newcommand{\Tab}[1]{Tab. \ref{#1}}

\title{\LARGE \bf
TextSLAM: Visual SLAM with Planar Text Features
}

\author{Boying Li, Danping Zou$^{*}$, Daniele Sartori, Ling Pei, and Wenxian Yu
\thanks{This work was  supported by the Grant  61405180104 from Chinese government. All the authors are with Shanghai Key Laboratory of Navigation and Location-based Service, Shanghai Jiao Tong University.} \thanks{ $^*$Corresponding author: Danping Zou ({\tt\small dpzou@sjtu.edu.cn)}}
}

\begin{document}

\maketitle
\thispagestyle{empty}
\pagestyle{empty}

\begin{abstract}

We propose to integrate text objects in man-made scenes tightly into the visual SLAM pipeline. The key idea of our novel text-based visual SLAM is to treat each detected text as a planar feature which is rich of textures and semantic meanings. The text feature is compactly represented by three parameters and integrated into visual SLAM by adopting the illumination-invariant photometric error. We also describe important details involved in implementing a full pipeline of text-based visual SLAM. To our best knowledge, this is the first visual SLAM method tightly coupled with the text features. We tested our method in both indoor and outdoor environments. The results show that with text features, the visual SLAM system becomes more robust and produces much more accurate 3D text maps that could be useful for navigation and scene understanding in robotic or augmented reality applications.

\end{abstract}

\section{INTRODUCTION}

	Visual SLAM is an important technique of ego-motion estimation and scene perception, which has been widely used in navigation for drones \cite{heng2014autonomous}, ground vehicles or self-driving cars \cite{lategahn2011visual}, and augmented reality applications \cite{chekhlov2007ninja}. 
A typical visual SLAM algorithm extracts point features \cite{mur2015orb,davison2003real} from images for pose estimation and mapping. Recent methods \cite{forster2014svo} \cite{engel2018direct} even directly operate on pixels. However, it is well known that incorporating high-level features like lines 
\cite{zhou2015structslam}, or even surfaces 
\cite{trevor2012planar} in the visual SLAM system will lead to better performance with fewer parameters.

One type of object surrounding us that can be used as high-level features is text. Text labels in daily scenes offer rich information for navigation. They can help us recognize landmarks, navigate in complex environments, and guide us to the destination.  Text extraction and recognition have been developing fast in these days \cite{zhu2016scene} \cite{yin2016text} because of the boom of the deep neural networks and the emergence of huge text datasets such as COCO-Text\cite{veit2016coco}, DOST\cite{iwamura2016downtown}, and ICDAR\cite{karatzas2015icdar}. One question raises whether texts can be integrated into a visual SLAM system to not only yield better performance but also generate high-quality 3D text maps that could be useful for navigation and scene understanding.
	
	\begin{figure}[t!]		
		\centering  
		\includegraphics[width=0.49\textwidth]{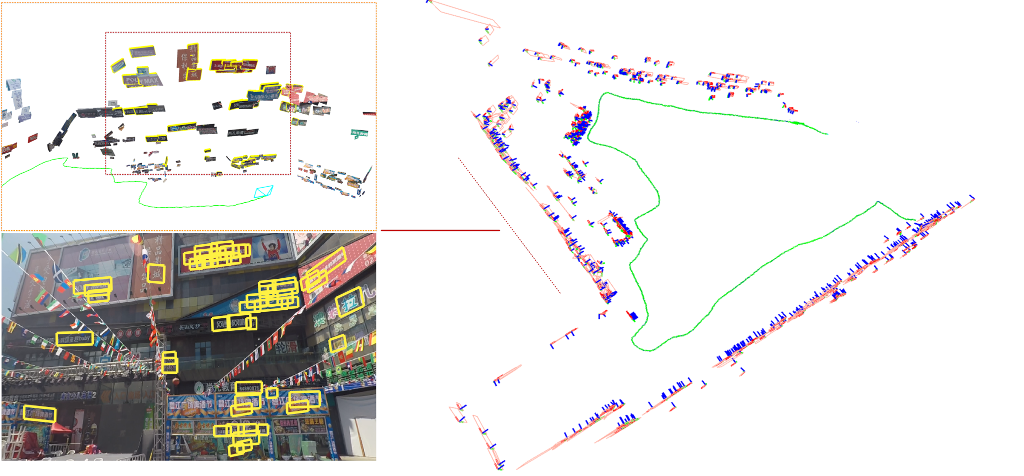}\\ 
		\caption{TextSLAM in a shopping mall.  \textbf{Left}: Detected texts in the images (in yellow rectangles) and the zoomed-in view of 3D text map. \textbf{Right}: 3D text maps and camera trajectory in top-down view. The text objects are illustrated in pink boxes and their normal directions are shown in blue.}  
		\label{fig:textEx}  
	\end{figure}
	There are several attempts towards text-aided navigation and location in recent years. 
	A navigation system \cite{rong2016guided, li2019vision} for blind people, assisted with text entities, is built upon the visual-inertial SLAM system shipped on the Google Tango tablet. 
	Similarly, Wang et al. proposed a method to extract text features \cite{wang2015bridging}, which are then used for fusion with Tango's SLAM outputs to facilitate closing loops. 
	The aforementioned works have shown the great potential of integrating text features with existing SLAM systems, though in a loosely coupled manner. 
	It is worth further investigation into a tightly coupled approach 
by putting texts into the SLAM pipeline instead of treating the existing SLAM system as a black box.
	
	We propose a novel visual SLAM method tightly coupled with text features. 
	Our basic motivation is that texts are usually planar and texture-rich patch features: Texts we spotted in our daily life are mostly planar-like regions, at least for a single word or character if not the whole sentence. The rich pattern of a text entity makes the text object a naturally good feature for tracking and localization.
	It is demonstrated in our work that through fully exploring those characteristics of text features, we can improve the overall performance of the  SLAM system, including the quality of both localization and semantic map generation. The main technical contributions of this paper includes:
	
	1) A novel three-variable parameterization for text features is proposed. The parameterization is compact and allows instantaneous initialization of text features with small motion parallaxes.
	
	2) The text feature is integrated in the visual SLAM system by adopting the photometric error measured by normalized sum of squared differences. Such photometric error is robust to illumination changes and blurry images caused by quick camera motions. Tracking and mapping of text features are done by minimizing the photometric errors without extra data association processes.
	
	3) We present details for implementing such a text-based SLAM system, including initialization and update of text features, text-based camera pose tracking and back-end optimization. To our best knowledge, this is the first visual SLAM method with text features tightly integrated. 
	
We conduct experiments in both indoor and outdoor environments. The
 results show our novel text-based SLAM method achieves better accuracy and robustness, and produces much better 3D text maps than does the baseline approach. Such semantically meaningful maps will benefit navigation in man made environments.

\section{RELATED WORK}

\paragraph{Planar features}
	Planar features have been studied in visual SLAM community since the early stage. In early works \cite{chekhlov2007ninja}\cite{gee2007discovering}\cite{gee2008discovering}, the planes in the scene were detected by RANSAC \cite{yang2010plane} among estimated 3D points and employed as novel features to replace points in the state. Much fewer parameters are required to represent the world using planes instead of points, hence reducing the computational cost significantly. Those works show that planar features improve both accuracy and robustness of a visual SLAM system.
However, existing methods require 3D information to discover the planar features, usually using a RGB-D camera \cite{sturm2012benchmark}\cite{kim2018linear}. This becomes difficult using only image sensors. An interesting idea \cite{molton2004locally} is to assume each image patch surrounded the detected feature point as one observation of a locally planar surface. The assumption however seldom holds in realistic scenes, as feature points might be extracted from anywhere in the scene.
	Nevertheless, texts in realistic scenes are mostly located on planar surfaces, though sometime only locally. Therefore texts are naturally good planar features that can be easily detected by text detectors.
	
\paragraph{Object features}

Integrating object-level features into visual SLAM systems has been receiving increasing interest in recent years \cite{salas2013slam++}\cite{galvez2016real}\cite{civera2011towards}.
	Existing methods however require pre-scanned 3D models to precisely fit the observation on the image. Though recent work \cite{mccormac2018fusion++} attempted to reconstruct the 3D model of objects online with a depth camera, it is still difficult to be generalized to unseen objects with only image sensors. Another solution is adopting 3D bounding boxes \cite{yang2019cubeslam}\cite{yang2019monocular} or quadrics \cite{nicholson2018quadricslam} to approximate generic objects. This however suffers from loss of accuracy. 	Unlike generic objects, the geometry of text objects is simple. As aforementioned, text objects can be treated as locally planar features and also contain a rich amount of semantic information about the environment. It is hence worth investigating how to design a SLAM system based on the text objects. 
	
\paragraph{Text-aided navigation}
	Text is a naturally good visual fiducial marker or optical positioning to assistant navigation \cite{houben2016joint}\cite{tapu2017computer} yet the technique is still under development \cite{houben2016joint}. Several loosely coupled text-aided navigation methods have shown a great potential for perception, navigation, and human-computer interaction.
	The early work \cite{tomono2000mobile} used the room number as a  guidance for robot to autonomously move in the office-like scenes. 
	The authors \cite{case2011autonomous} annotated the text labels on the existing map generated by a laser SLAM system to help robots understand each named location. 
	With the prior knowledge of a comprehensive map and the compass information, the authors  \cite{radwan2016you} extracted 2D text information from the observations to assist localization.  
	In the work \cite{wang2015bridging}, the spatial-level feature named 'junction' was extracted from text objects, and then combined with the location and mapping output of Google Tango's SLAM system at the stage of loop closing. 
	The authors present a text spotting method \cite{rong2016guided} for the assistant navigation system relying the Tango's SLAM system.
	Similarly, with the SLAM system of Tango, a mobile solution \cite{li2019vision} of assistant navigation system combines various sources, such as text
recognition results and speech-audio interaction, for blind and visually impaired people to travel indoor independently.	
	
	Existing text-aided methods regard SLAM systems (either vision-based or laser-based) as a black box and take less attention on the accuracy of 3D text maps. 
	By contrast, the proposed method integrates the text objects tightly into the SLAM system to facilitate both camera tracking and mapping, and focuses on generating high accurate text map that can be used for future pose estimation and loop detection.

\section{Text features}
\subsection{Parameterization} 

\begin{figure}[!tbp]		
		\centering  
		\includegraphics[width=0.35\textwidth]{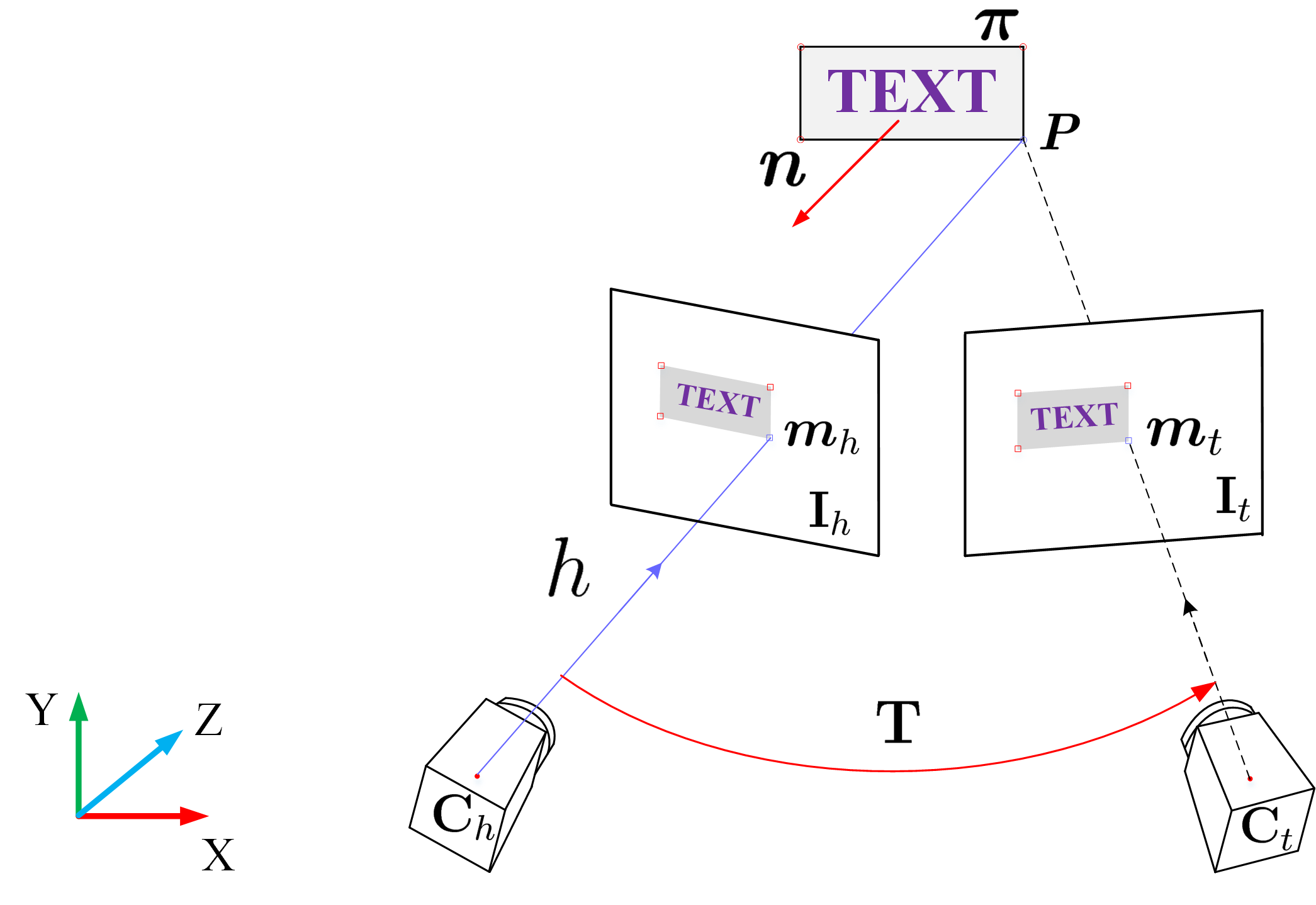}\\  
		\caption{A text object is compactly parameterized by $\bv{\theta}$. The inverse depth $\rho$ of a text point $\bv{p}$ can be computed by $\rho = 1/h = \bv{\theta}^\vts \tilde{\bv{m}}$ and its projection onto the target view $C_t$ is a homography transform with respect to the relative pose $\ba{T}$ between the two views.} 
		\label{fig:textParam}  
	\end{figure}

	As discussed previously, most text objects can be regarded as planar and bounded patches. Each text patch (usually enclosed by a bounding box) is anchored to the camera frame, named as the \emph{host frame}, when it is firstly detected on the image as shown in \Fig{fig:textParam}.
	Expressed in the coordinate system of the host frame, the plane where the text patch lies is described by the equation 
$ \bv{n}^\vts\bv{p}+d = 0,
$	where $\bv{n}=(n_1,n_2,n_3)^\vts\in\mathbb{R}^{3}$ is the normal of the plane and $d \in \mathbb{R}$ is related to the distance from the plane to the origin of the host frame; $\bv{p}\in\mathbb{R}^{3}$ represents the 3D point on the plane.
	
	A straightforward parameterization of a text plane could be directly using the four parameters $(n_1,n_2,n_3,d)$ of the plane equation. But this is a over parameterization that leads to rank deficient in the nonlinear least-squares optimization. We propose to use a compact parameterization that contains only three parameters.
	\beq
	\bv{\theta}=(\theta_1,\theta_2,\theta_3)^\vts = -\bv{n}/d.
	\eeq  We'll show that this parameterization is closely related to the inverse depth of the 3D point on the text plane.

	Within the host frame, each 3D point $\bv{p} \in \mathbb{R}^{3}$ observed on the image is able to be represented by its normalized image coordinates $\bv{m}=(u,v)^\vts$ and its inverse depth $\rho = 1/h$ 
	, where the depth $h$ is the distance between this 3D point and the camera center. The 3D coordinates of this 3D point are computed as $\bv{p} = (uh,vh,h)^\vts = h\tilde{\bv{m}}$, where $\tilde{\bv{m}}$ denotes the homogeneous coordinates of $\bv{m}$. If the 3D point locates on the text plane, we have $h \cdot\bv{n}^\vts\tilde{\bv{m}}+d=0$.	
%
	The inverse depth $\rho$ of this 3D point is then computed as
	\beq
	\rho = 1/h = - \bv{n}^\vts/d\,\,\tilde{\bv{m}} = \bv{\theta}^\vts 	
	\tilde{\bv{m}}.
	\label{eq:invdepth_01}
	\eeq
	That is, we can use a simple dot product to quickly infer the inverse depth of a text point from its 2D coordinates, given the text parameters $\bv{\theta}$.
	
	On the other hand, if we have at least three points on the text patch (for example, three corners of the bounding box), with their inverse depths, we can immediately obtain the text parameters by solving
	\beq
	\bmat
	\tilde{\bv{m}}^\vts_1\\
	\vdots\\
	\tilde{\bv{m}}^\vts_n
	\emat \bv{\theta} = 
	\bmat
	\rho_1\\
	\vdots\\
	\rho_n
	\emat, n \ge 3.
	\label{eq:param_from_invdepth}
	\eeq
	This allows us to quickly initialize the text parameters from the depth value of three corners of the text bounding box. 
	
To fully describe a text object, properties such as the boundary of the text object and pixel values are also kept in our system. Those information can be acquired from a text detector as we'll described later.

\subsection{Projection of the 3D text object onto a target image}

	Here we describe how to project a 3D text object anchored at a host frame onto the image related to the \emph{target frame}. 
	Let $\ba{T}_{h} ,\ba{T}_{t}\in SE(3)$ represent the transformations from the host frame and the target frame to the world frame respectively. The transformation from the host frame to the target frame is $\mathbf{T} =\mathbf{T}_{t}^{-1} \mathbf{T}_{h}$. We let $\ba{R},\bv{t}$ be the rotation and translation of $\ba{T}$. 	Given the text parameters $\bv{\theta}$ and the observed image point $\bv{m}$ (with the homogeneous coordinates $\tilde{\bv{m}}$) in the host frame, the 3D coordinates of point $\bv{p}$ are :	\beq
	\bv{p} = \tilde{\bv{m}}/\rho=\tilde{\bv{m}}/(\bv{\theta}^\vts \tilde{\bv{m}}).
	\eeq
	The point is then transformed into the target frame. Let the transformed point be $\bv{p}'$. We have $
	\bv{p}' \sim \ba{R} \tilde{\bv{m}} + \bv{t}\, \tilde{\bv{m}}^\vts\bv{\theta}.$
	Here notation $\sim$ means equivalence up to a scale. Using the pinhole camera model, the coordinates of image point $\bv{m}'= (u',v')^\vts$ of $\bv{p}'$ in the target frame are computed as
	\beq
	\begin{array}{rl}
		u'&=(\bv{r}_1\tilde{\bv{m}}+t_1 \tilde{\bv{m}}^\vts\bv{\theta})/(\bv{r}_3\tilde{\bv{m}}+t_3 \tilde{\bv{m}}^\vts\bv{\theta})\\
		v'&=(\bv{r}_2\tilde{\bv{m}}+t_2 \tilde{\bv{m}}^\vts\bv{\theta})/(\bv{r}_3\tilde{\bv{m}}+t_3 \tilde{\bv{m}}^\vts\bv{\theta})
	\end{array},
	\label{eq:homography}
	\eeq
	where $\bv{r}_1,\bv{r}_2,\bv{r}_3$ are the row vectors of $\ba{R}$ and 
	$\bv{t}=(t_1,t_2,t_3)^\vts$. Note that \Eq{eq:homography} is in fact the homography transformation from $\bv{m}$ to $\bv{m}'$, where the homograpny matrix is defined as  $
	\ba{H} \sim \ba{R}+\bv{t}\bv{\theta}^\vts.
$
	Hence, the whole process of projecting a 3D text object on the image plane of a target frame can be described as a homography mapping
	\beq
	\bv{m}' = \bv{h}(\bv{m},\ba{T}_{h},\ba{T}_t,\bv{\theta}).
	\label{eq:text_prediction}
	\eeq
	
\subsection{Photometric error for text objects}
		Photometric error is used to compare the projected text object and the observed one on the image. This is done by pixel-wise comparison  and similar to the direct approaches 
\cite{engel2018direct} which have been shown to be accurate and robust without finding corresponding feature points explicitly. 
	The biggest issue of using photometric error is to handle the intensity changes. Existing work \cite{engel2018direct} adopts an affine model to address intensity changes, but it requires extra parameters involved in optimization and sophisticated photometric calibration to guarantee performance. 
	
	We choose to use zero mean normalized cross-correlation (ZNCC) as the matching cost to handle illumination changes. Let $\Omega$ be the set of pixels within the text region, and $\bv{m} \in \Omega$ be a text pixel. The normalized intensities for text pixels are:
$
	\tilde{{I}}(\bv{m}) = ({I}(\bv{m}) - \bar{{I}}_\Omega)/(\sigma_\Omega \sqrt{N}),
$
	where $\bar{{I}}_{\Omega}$ and $\sigma_{\Omega}$ stand for the average intensity and the standard deviation of the pixels in $\Omega$, and $N$ is the number of pixels. 
	 The text patch in the host frame and the predicted one in the target frame \Eq{eq:text_prediction} are then compared by :
\beq
	ZNCC({{I}}_h,{{I}}_t) = \sum_{\bv{m}\in\Omega} \tilde{{I}}_h(\bv{m}) \tilde{{I}}_t(\bv{m}').
	\label{eq:zncc}
\eeq
	The ZNCC cost  is between $-1$ and $1$. The larger ZNCC cost indicates the two patches are more similar. However, it is difficult to directly use 
	the ZNCC cost in visual SLAM, since it can not be formulated as a nonlinear least squares problem. We therefore adopt a variant form of ZNCC as the cost function
	\beq
	E({{I}}_h,{{I}}_t) = \sum_{\bv{m}\in\Omega} (\tilde{{I}}_h(\bv{m})-\tilde{{I}}_t(\bv{m}'))^2.
	\label{eq:norm_ssd}
	\eeq
 Though the cost function is similar to the SSD (Sum of Squared Difference) cost, it contains additional normalization process to ensure the robustness to illumination changes. If we expand this cost function as : 
	\beq
\sum_{\bv{m}\in\Omega}(\tilde{{I}}_h(\bv{m})^2 +\tilde{{I}}_t(\bv{m}')^2) - 2\sum_{\bv{m}\in\Omega}\tilde{{I}}_h(\bv{m}) \tilde{{I}}_t(\bv{m}'), 
	\eeq
	we discover that minimizing this cost function is equivalent to maximizing the ZNCC cost, because  $\sum\tilde{{I}}_h(\bv{m})^2 = 1$ and $\sum\tilde{{I}}_t(\bv{m}')^2 = 1$. The photometric error of a text object $\pi$ with respect to the target frame $t$ is defined as :
	\beq
	E^{\pi,t}_{photo} = \sum_{\bv{m}\in\Omega^\pi} \phi((\tilde{{I}}_h(\bv{m})-\tilde{{I}}_t(\bv{h}(\bv{m},\ba{T}_h, \ba{T}_t, \bv{\theta}^\pi)))^2),
	\label{eq:photometric_error}
	\eeq
	where $\phi(\cdot)$ is the Huber loss function to handle possible outliers. Here, we use $\Omega^\pi$ to represent the text region on the image plane in the host frame. As we'll describe later, to make the computation faster, we do not use all the pixels within the text region, instead select only some of them as the reference pixels to compute the photometric error. 
	
\section{TEXTSLAM SYSTEM}

\begin{figure}[!tbp]		
		\centering  
		\includegraphics[width=0.48\textwidth]{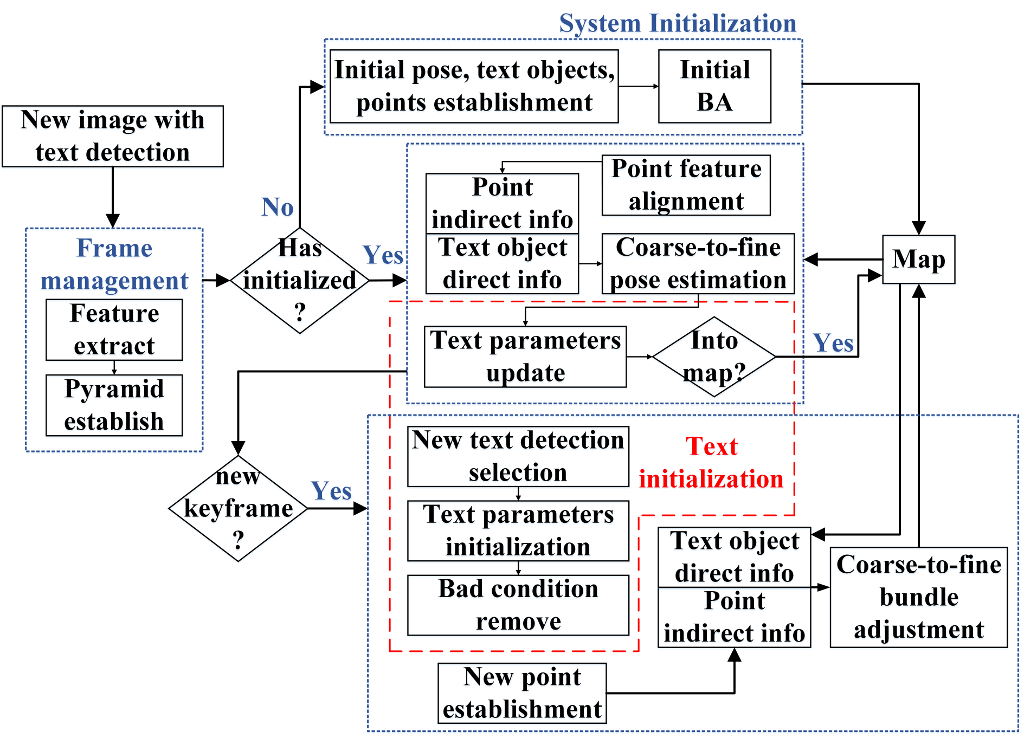}\\  
		\caption{An overview of TextSLAM system.}  
		\label{fig:flowchart}  
	\end{figure}

	Our TextSLAM system is built upon the basic system using point features and adopts the keyframe-based framework to integrate the text features tightly. 
The mixture of point features and text features allow our system to work properly even in the scenes without text labels. 
	Fig. \ref{fig:flowchart} illustrates the flowchart of TextSLAM. we'll detail the key components in the following sections.

\subsection{Initialization of text objects}

	Text objects are extracted on the image whenever a new image has been acquired. 
The deep learning technique has largely accelerated the text extraction development in recent years \cite{liu2018fots, he2017single, busta2017deep, he2017deep}.
The text detector named EAST \cite{zhou2017east} is used to extract objects in our implementation, but other text detectors may also be used, because our system does not rely on particular text detectors.
	Some examples of text extraction are demonstrated in \Fig{fig:textEx}. The outputs are arbitrary-orientation quadrilaterals enclosing the text regions. Once a text object has been extracted, 
	we detect FAST \cite{rosten2006machine} features within the text region and track them via Kanade-Lucas-Tomasi (KLT) \cite{shi1993good} until the next key frame. Then the parameters of the text object are initialized from the tracked points. Let $\bv{m}_i  \leftrightarrow \bv{m}'_i$ be the corresponding points in both views, and $\ba{R},\bv{t}$ be the transformation between the two frames. From \Eq{eq:homography}, we have
	\beq
	[\tilde{\bv{m}}'_i]_\times \bv{t}\tilde{\bv{m}}^\vts_i \bv{\theta} =  -[\tilde{\bv{m}}'_i]_\times\ba{R}\tilde{\bv{m}}_i,
	\eeq
	where $\tilde{\bv{m}}_i$ and $\tilde{\bv{m}}'_i$ are the homogeneous coordinates of $\bv{m}_i$ and $\bv{m}'_i$. Note that the rank of the matrix on the left hand side  is one. It requires at least three pair of corresponding points to solve $\bv{\theta}$. The text parameter is futher refined by minimizing the photometric error as defined in \Eq{eq:photometric_error}. 
	
	After initialization, we keep the text quadrilateral with the text object. The four corners can be projected onto other views  \Eq{eq:text_prediction} to predict the appearance. Note that a text object can only be initialized when its quadrilateral in the current view is not intersected with existing text objects or partially out of the image. Otherwise they are rejected as 'bad initialization'. 
	The newly initialized text objects are kept being updated in the following frames. They are inserted into the map only if the text object has been observed in at least  $n_{min}$ ($5$ in our implementation) frames.

\subsection{Camera pose estimation with text objects}
Both points and text objects in the map are involved in camera pose estimation. The camera pose estimation is to minimize the following cost function
	\beq
	E(\ba{T}_t) = E_{point}(\ba{T}_t) + \lambda_w E_{text}(\ba{T}_t),
	\label{eq:pose_estimation_cost}
	\eeq
	where $\ba{T}_t \in SE(3)$ represents the current camera pose. $E_{point}$ and $E_{text}$ come from the feature points and the text objects respectively. Note that $E_{point}$ consists of geometric errors, or reprojection errors, but $E_{text}$ contains only photometric errors 
	\beq
	E_{text} = \sum_{\pi} E^{\pi,t}_{photo}.
	\eeq The trade-off between them needs to be regulated by the weight $\lambda_w$ since they are in different units (position difference vs intensity difference). 
	
	The weight $\lambda_w$ is computed as $\lambda_w = \sigma_{rep}/\sigma_{photo}$. $\sigma_{rep}$ represents the standard deviation of the reprojection error of a pair of corresponding points (in both $x$ and $y$ directions) and $\sigma_{photo}$ represents the standard deviation of the photometric error of a text object as defined in \Eq{eq:norm_ssd}. Those standard deviations can be acquired through a small set of training data  (given corresponding points and text patches).
	
	Optimization of the cost function \Eq{eq:pose_estimation_cost} is a nonlinear least squares problem. As the photometric cost $E_{text}$ is highly nonlinear, it requires a good initial guess of $\ba{T}_t$ to avoid being trapped in a local minimum. We firstly use a constant velocity model to predict the camera pose. Based on this prediction, we then apply a coarse-to-fine method for the optimization, and the camera pose is estimated by minimizing \Eq{eq:pose_estimation_cost} iteratively.

\subsection{Bundle Adjustment with text objects}

	We apply bundle adjustment from time to time in a local window of key frames similar to \cite{mur2015orb}. The cost function of bundle adjustment consists the point part and the text part :
	\beq
	E(\bv{x}) = E_{point}(\bv{x}) + \lambda_w E_{text}(\bv{x}).
	\label{eq:ba_cost}
	\eeq
	The cost function resembles that of camera pose estimation while involves more parameters to be optimized. The variable $\bv{x}$ include the camera poses of key frames in the local window, the 3D coordinates of point features, and the text parameters.
	 We also adopt a coarse-to-fine method to optimize \Eq{eq:ba_cost} as in camera pose estimation.

	Though we may use all the pixels within the text region to evaluate the photometric errors, a more efficient way is to use a small part of them. The representative pixels  are selected with the minimum number of $15$ by FAST \cite{rosten2006machine}, and further applied to evaluate photometric errors. 
	
\section{Experiment}
\subsection{Data collection}
\begin{figure}[]		
		\centering  
		\includegraphics[width=0.48\textwidth]{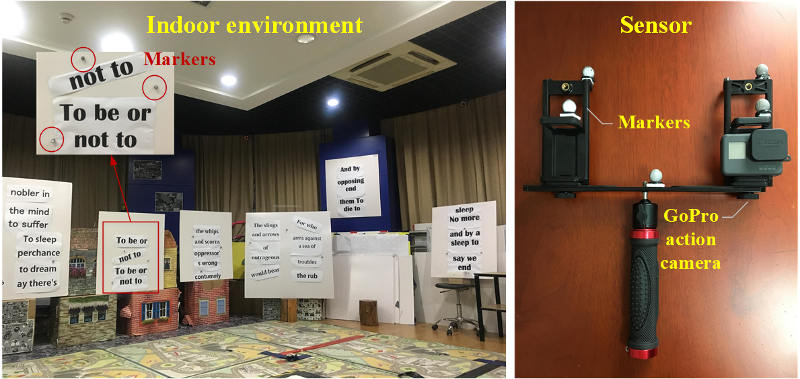}\\  
		\caption{The indoor test scene is shown on the left. The data collection device equipped with the  GoPro camera, is presented on the right.}  
		\label{fig:ExpSensor}  
	\end{figure}

	We collected a set of image sequences for evaluation in both indoor and outdoor scenes.
	\Fig{fig:ExpSensor} shows our device for the data collection. It consists of the hand-held camera and optical markers for the acquisition of ground truth trajectories.
All image sequences were resized to $640\times 480$ for tests. 
	
\subsection{Indoor scene with ground truth}
	The indoor  environment, with text labels randomly placed, is shown in \Fig{fig:ExpSensor}, which is equipped with a motion capture system to obtain the ground truth trajectories of millimeter accuracy within an area of $4 m \times 4 m$.
Three methods were evaluated : our text-based method, our point-only method, and ORB-SLAM \cite{mur2015orb} where loop closing was disabled for fair comparison.
	
\paragraph{Trajectory estimation}	

	\begin{table}
		\caption{Table I: Localization performance. RPE (0.1 m) and APE (m)}
		\begin{center}
		\begin{tabular}{ccccccc}
			\hline
			\multicolumn{1}{c}{\multirow{2}{*}{\textbf{Seq.}}} & \multicolumn{2}{c}{\textbf{ORB-SLAM}}             & \multicolumn{2}{c}{\textbf{Point-only}}     & \multicolumn{2}{c}{\textbf{TextSLAM}}             \\ \cline{2-7} 
			\multicolumn{1}{c}{}                                  & \multicolumn{1}{c}{RPE} & \multicolumn{1}{c}{APE} & \multicolumn{1}{c}{RPE} & \multicolumn{1}{c}{APE} & \multicolumn{1}{c}{RPE} & \multicolumn{1}{c}{APE} \\ \hline
			Indoor\_01       &  0.342     &  0.161     &  0.192     &  0.223     &  0.188     &  0.196    \\ 
			Indoor\_02       &  0.300     &  0.150     &  0.189     &  0.164     &  0.187     &  0.150    \\ 
			Indoor\_03       &  0.228     &  0.144     &  0.207     &  0.166     &  0.209     &  0.155    \\ 
			Indoor\_04       &  0.188     &  0.145     &   --       &   --       &  0.173     &  0.171    \\ 
			Indoor\_05       &  0.277     &  0.120     &  0.172     &  0.160     &  0.174     &  0.161    \\ \hline
		\end{tabular}\\
\end{center}
		\emph{The bar '--' indicates the algorithm fails to finish the whole trajectory.}
		\label{tab:generalIndoor}
	\end{table}
We present both the relative pose error (RPE) and the absolute pose error (APE) in \Tab{tab:generalIndoor}. Note that the errors of those methods are very close to each other. The reason must be the test scene is very small and highly textured, in which using only feature points should work well.
ORB-SLAM slightly outperforms both of our methods, which is not surprising because ORB-SLAM adopts a sophisticated map reuse mechanism based on the covisibility pose graph. Our methods currently are odometry systems virtually. Nevertheless, we still observe a performance gain in using text features compared with our point-only implementation.

\begin{figure}		
		\centering  
		\includegraphics[width=0.49\textwidth]{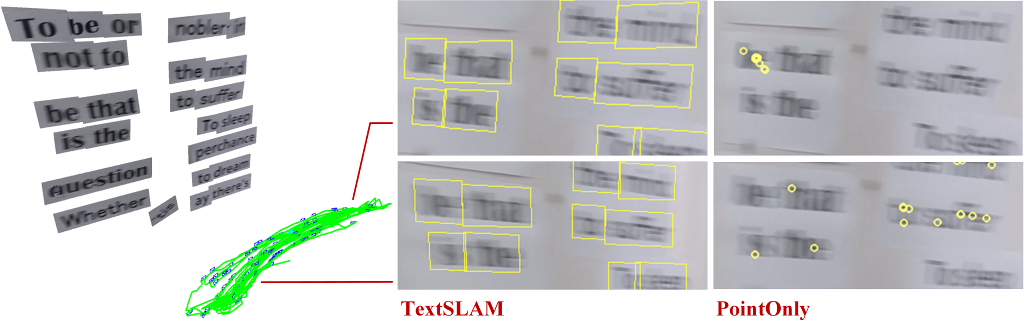}\\  
		\caption{TextSLAM is robust to blurry images caused by rapid camera motions. The estimated 3D text map and  camera trajectory of TextSLAM are shown on the left. By contrast, point-only method failed to track feature points on severely blurry images as shown on the right.}

		\label{fig:rapidData}  
	\end{figure}

We also evaluate the robustness of the proposed method under fast camera motion. The rapid motion causes severe image blur as shown in \Fig{fig:rapidData}, inducing the failure of our point-only method in all cases. ORB-SLAM failed only at one test. This is also because its well implemented relocalization mechanism. By contrast, our text-based method works well in those tests and performs much more accurate as shown in \Tab{tab:rapid}. This is largely due to our direct approach towards text objects using photometric errors.

	\begin{table}[]
		\caption{Table II: Indoor rapid performance. RPE (0.1 m) and APE (m)}
		\begin{center}
		\begin{tabular}{ccccccc}
			\hline
			\multirow{2}{*}{\textbf{Seq.}} & \multicolumn{2}{c}{\textbf{ORB-SLAM}} & \multicolumn{2}{c}{\textbf{Point-only}} & \multicolumn{2}{c}{\textbf{TextSLAM}} \\ \cline{2-7} 
			& RPE               & APE               & RPE                   & APE                   & RPE               & APE               \\ \hline
			Rapid\_01                        & --                & --                & --                    & --                    & 0.271            & 0.029            \\ 
			Rapid\_02                        & 0.367            & 0.063            & --                    & --                    & 0.322            & 0.031            \\ 
			Rapid\_03                        & 0.338            & 0.061            & --                    & --                    & 0.212            & 0.022            \\ \hline
		\end{tabular}\\
	\end{center}
		\emph{The bar '--' indicates the algorithm fails to finish the whole trajectory.}
		\label{tab:rapid}
	\end{table}

		\begin{figure*}[!ht]		
	 	\centering  
	 	\includegraphics[width=0.98\textwidth]{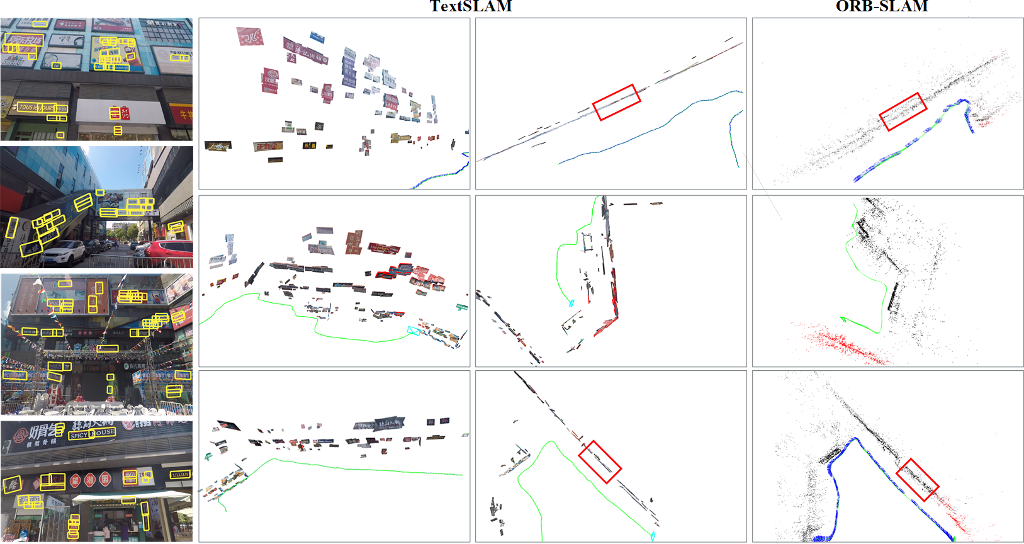}\\ 
	 	\caption{Real world tests in a shopping center. The text detection results are shown in the first column. Three typical locations are shown and enlarged in each row. The second and third columns show the close-up and top-down views. For comparison, the ORB-SLAM performance of the same locations are also presented in the fourth column. We can observe the noisy point clouds visually, as enclosed in red rectangles.} 
	 	\label{fig:BiJiangMap}
	\end{figure*}
	
			\begin{figure}[!h]		
		\centering  
		\includegraphics[width=0.45\textwidth]{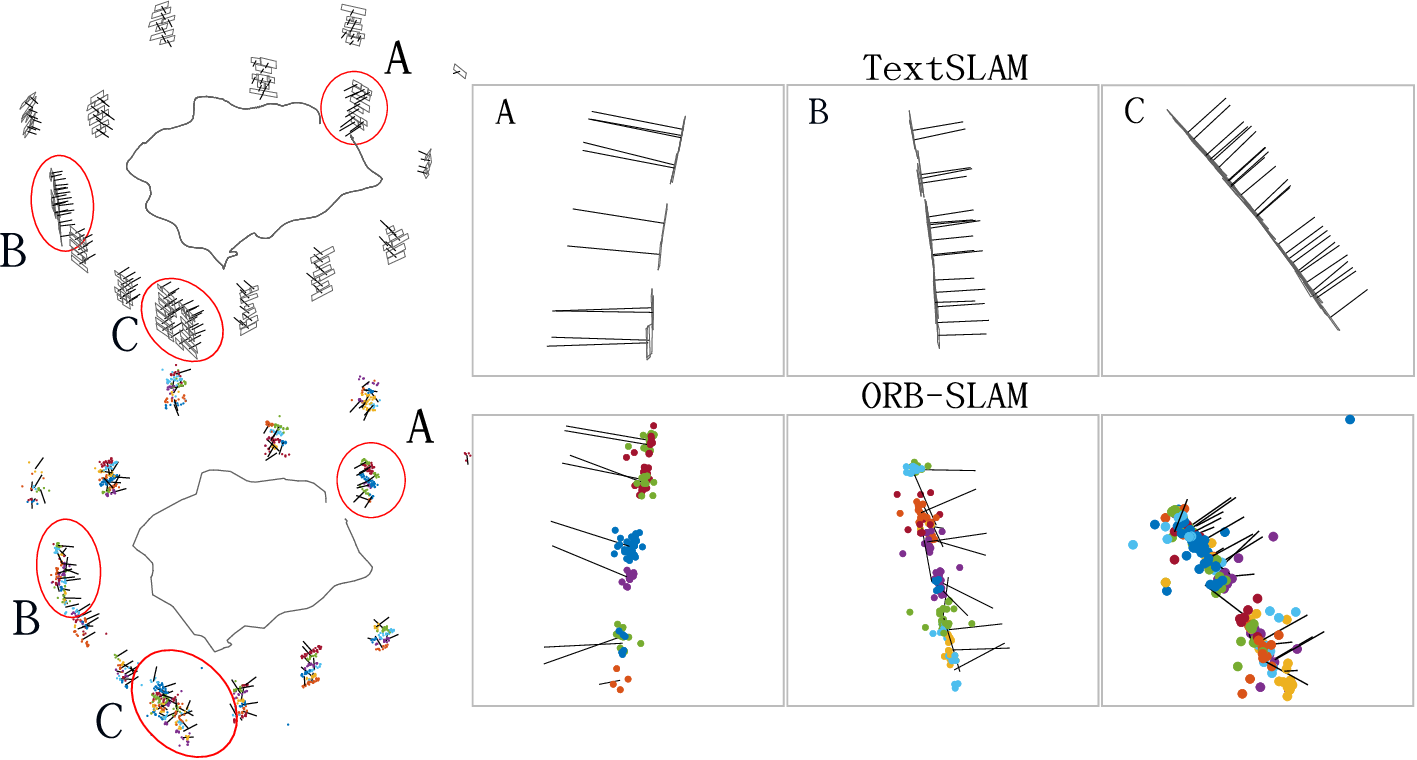}\\  
		\caption{
		Though RANSAC was adopted, plane fitting on the point clouds from ORB-SLAM still produced noisy results (as shown in the bottom row). TextSLAM avoids such problem by matching or tracking a text object as a whole by using photometric
errors.}
		
		\label{fig:TextORBMap}  
	\end{figure}

	\begin{figure}[!h]
		\centering  
		\includegraphics[width=0.45\textwidth]{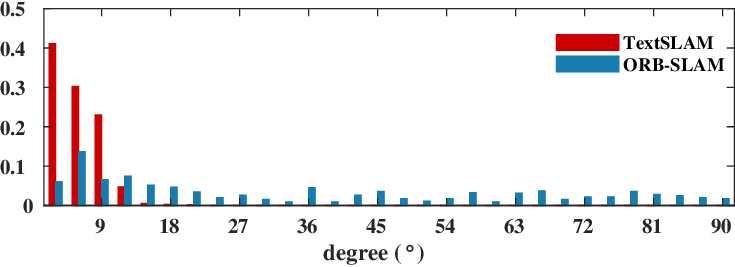}\\  
		\caption{The statistic distribution of the angular errors.
		The results of TextSLAM and ORB-SLAM are illustrated in red and blue, respectively.}
		\label{fig:Neval_Text}  
	\end{figure}
	
\paragraph{3D text maps}
	We use the angular error of each estimated text plane and visually inspection to evaluate the mapping performance. 
	The angular error measures the difference between the estimated normal $\bv{n}_{t}$ of the text plane and the ground truth $\bv{n}_{gt}$, namely $\alpha = \arccos(|\bv{n}_{t}^\vts\bv{n}_{gt}|/\|\bv{n}_{t}\|\|\bv{n}_{gt}\|)$. $\bv{n}_{gt}$ was acquired by a few optical markers on the text plane as shown in \Fig{fig:ExpSensor}. 
	Since no visual SLAM system generates text maps directly available for the evaluation, we implemented a loosely-coupled system based on ORB-SLAM for comparison by fitting the text plane from the 3D text points using three-point RANSAC.
 
	The statistic of angular errors are presented in \Fig{fig:Neval_Text} and one of the mapping results is visually presented in \Fig{fig:TextORBMap}.
	The results demonstrate that the 3D text map produced by TextSLAM is substantially better than that of the plane fitting approach based on ORB-SLAM. 
	The reason is that the point clouds generated by ORB-SLAM are in fact noisy as shown in \Fig{fig:TextORBMap}. 
We carefully checked what causes those noisy points and speculate that it may be caused by incorrect feature location or correspondences in the images.
	 
\subsection{Real-world tests}
	
	The outdoor experiment is in a commercial center as shown in the first column in \Fig{fig:BiJiangMap}. 
This daily environment is full of various challenges, including text objects with various sizes, fonts, backgrounds and languages, the complex occlusion, the reflection of glass and the dynamic pedestrians.	

	Since it is difficult to acquire the ground truth for either the camera trajectory or the 3D text map, we present only visual results to show the efficacy of our method. 
	The second and three columns in \Fig{fig:BiJiangMap} illustrate reconstructed 3D text labels and the estimate trajectories. A side-by-side comparison with ORB-SLAM from the top-down view is also presented on the last column.
	
The deep learning text detector sometimes produces false detections, the text object initialization (introduced in SectionIV-A) filters out bad text objects automatically before insertion to the map. Our TextSLAM also works properly when no text labels found, such as the parking lot of the commercial center.

	Though no ground truth is available, we can still observe the noisy point clouds in the ORB-SLAM results. 
	As highlighted by rectangles in \Fig{fig:BiJiangMap}, the text maps better reveal the planar structures of the scene than the noisy point clouds from ORB-SLAM.
	As aforementioned, the noisy points are caused by  either incorrect
feature matching or detection,  preventing ORB-SLAM from acquiring accurate 3D text maps.

\section{Conclusion \& Future work}
	We present a novel visual SLAM method tightly coupled with the planar text features. Experiments have been conducted in both  artificial indoor cases with ground truth and real world scenes. The results show that our text-based SLAM method performs better than point-only SLAM method does, especially in blurry video sequences caused by rapid camera motions. However, the localization performance gain is not as large as we expected in the indoor tests, even when point-based methods generate very noisy point clouds. This could be that camera pose estimation is less sensitive to noisy points because robust approaches are usually involved. Nevertheless, the results demonstrate that our method generates much more accurate 3D text maps than  the loosely-coupled method based on the state-of-the-art visual SLAM system. Future work includes incorporating  text semantics into our system and acquiring real world datasets with highly accurate ground truth for quantitative evaluation.


\bibliographystyle{IEEEtran}
\bibliography{IEEEabrv,ReferenceTex}
\end{document}